\documentclass[submission,copyright,creativecommons]{eptcs}
\providecommand{\event}{FMAS 2023} % Name of the event you are submitting to

\usepackage{iftex}
\usepackage{graphicx}
\usepackage[utf8]{inputenc}
\usepackage[T1]{fontenc}

\usepackage[english]{babel}

\usepackage{siunitx}
\usepackage{wrapfig}
\BeforeBeginEnvironment{wrapfigure}{\setlength\intextsep{0pt}}
\usepackage{microtype}
\usepackage{framed}

\usepackage{subfig}
\usepackage{graphicx}

\usepackage{url}
\usepackage{fancyvrb}

\usepackage[babel]{csquotes}

\usepackage{todonotes}
\presetkeys{todonotes}{inline}{}

\usepackage{bsymb}
\usepackage[misc]{ifsym}

\usepackage{amsmath}
\usepackage{amsfonts}
\usepackage{amssymb}

\usepackage{hyperref}
\usepackage[capitalise,noabbrev]{cleveref}

\usepackage{listings}
\AtBeginDocument{}

\usepackage{comment}

\usepackage{booktabs}
\usepackage{tabularx}
\usepackage{siunitx}
\usepackage{wrapfig}
\BeforeBeginEnvironment{wrapfigure}{\setlength\intextsep{0pt}}
\usepackage{microtype}
\usepackage{bsymb}
\usepackage{framed}

\usepackage{xurl}
\usepackage{hyperref}
\usepackage[capitalise,noabbrev]{cleveref}

\usepackage{comment}

\newcommand{\prob}{{\sc ProB}}

\newcommand{\ignore}[1]{}

\usepackage{xcolor}
\definecolor{orcidlogocol}{HTML}{A6CE39}
\usetikzlibrary{svg.path}
\tikzset{
  orcidlogo/.pic={
    \fill[orcidlogocol] svg{M256,128c0,70.7-57.3,128-128,128C57.3,256,0,198.7,0,128C0,57.3,57.3,0,128,0C198.7,0,256,57.3,256,128z};
    \fill[white] svg{M86.3,186.2H70.9V79.1h15.4v48.4V186.2z}
    svg{M108.9,79.1h41.6c39.6,0,57,28.3,57,53.6c0,27.5-21.5,53.6-56.8,53.6h-41.8V79.1z M124.3,172.4h24.5c34.9,0,42.9-26.5,42.9-39.7c0-21.5-13.7-39.7-43.7-39.7h-23.7V172.4z}
    svg{M88.7,56.8c0,5.5-4.5,10.1-10.1,10.1c-5.6,0-10.1-4.6-10.1-10.1c0-5.6,4.5-10.1,10.1-10.1C84.2,46.7,88.7,51.3,88.7,56.8z};
  }
}
\newcommand{\orcidID}[1]{%
  \resizebox{8px}{8px}{
    \href{https://orcid.org/#1}{\tikz[yscale=-1,transform shape]{\pic{orcidlogo}}}}%
}

\usepackage{color}
\definecolor{lightgray}{rgb}{.9,.9,.9}
\definecolor{darkgray}{rgb}{.4,.4,.4}
\definecolor{purple}{rgb}{0.65, 0.12, 0.82}

\usepackage{makecell}

\ifpdf
  \usepackage{underscore}         % Only needed if you use pdflatex.
  \usepackage[T1]{fontenc}        % Recommended with pdflatex
\else
  \usepackage{breakurl}           % Not needed if you use pdflatex only.
\fi

\title{Certified Control for Train Sign Classification\thanks{This research is part of the KI-LOK project funded by the ``Bundesministerium f\"ur Wirtschaft und Energie''; grant \# 19/21007E.}}
\author{Jan Roßbach\orcidID{0009-0005-7725-9832}
  \institute{
    Heinrich-Heine-Universit\"{a}t D\"{u}sseldorf\\
    Mathematisch-Naturwissenschaftliche Fakultät\\
    Institut f\"{u}r Informatik}
  \email{jan.rossbach@uni-duesseldorf.de}
  \and
  Michael Leuschel\orcidID{0000-0002-4595-1518}
  \institute{
    Heinrich-Heine-Universit\"{a}t D\"{u}sseldorf\\
    Mathematisch-Naturwissenschaftliche Fakultät\\
    Institut f\"{u}r Informatik}
  \email{leuschel@uni-duesseldorf.de}
}
\def\titlerunning{Certified Control for Train Sign Classification}
\def\authorrunning{Jan Roßbach, Michael Leuschel}
\begin{document}
\maketitle

\begin{abstract}

  There is considerable industrial interest in integrating AI techniques
  into railway systems, notably for fully autonomous train systems.
  The KI-LOK research project is involved in developing new methods for
  certifying such AI-based systems.
  Here we explore the utility of a certified control architecture
  for a runtime monitor that prevents false positive detection of traffic signs
  in an AI-based perception system. The monitor uses classical computer vision algorithms to check
  if the signs -- detected by an AI object detection model -- fit
  predefined specifications. We provide such specifications for some critical
  signs and integrate a Python prototype of the monitor with
  a popular object detection model to measure relevant performance metrics on generated data.
  Our initial results are promising, achieving considerable precision gains with
  only minor recall reduction; however, further investigation into
  generalization possibilities will be necessary.

\end{abstract}

\section{Introduction and Motivation}

Artificial intelligence has been increasingly used in
various sectors, including transportation~\cite{RistiDurrant2021ARO}.
One particular area where artificial intelligence (AI) has gained attention is the development of
autonomous driving systems for railways~\cite{StandardisationConsiderations}.
The results already achieved in other transport sectors, mainly automotive, have
encouraged the development of AI in the railway industry~\cite{TANG2022103679}.

While this technology holds high economic interest, reliable certification methods are necessary to ensure
safe and regulated access to these innovations~\cite{StandardisationConsiderations}.
Traditional verification approaches such as formal methods have faced
difficulties in this area due to the opaque nature of AI, particularly
in computer vision where class definitions for classification tasks based on
raw pixel values have been considered challenging.

The KI-LOK~\footnote{\url{https://ki-lok.itpower.de}} research project addresses these challenges by developing
certification methodologies for autonomous AI-based railway systems.
As part of this, a case study~\cite{shuntingPaper} on train movements during
shunting movements is being analyzed. A formal B~\cite{bmethod} model has been
developed~\cite{shuntingPaper} to analyze the environment and ensure the
safety of the deterministic steering system through model checking with the
\prob~\cite{prob} model checker. The safety of the system was found to be
conditional on correct results from the AI-based perception system.
In this work, we attempt to move towards verification of part of this perception system
using a runtime monitor with a certified control~\cite{certcontrol}
architecture. This architecture reduces the part of the system requiring formal verification
compared to traditional monitor architectures putting a more formal analysis back into reach.
In particular, we focus on a subset of the train sign classification component.
It is responsible for detecting and classifying signs in the shunting yard to
ensure safe train movements. False recognition of a 'track-free' (Sh1) signal has
been determined to have safety implications.
We aim to significantly reduce or eliminate such false positives for some
of the most critical classes by defining a sign-specific ontology and checking
it at runtime. For this we introduce such a specification and show the
potential performance gains by evaluating a prototype implementation in Python on a custom dataset.

\section{Background and Related Work}
\label{sec:related}

The case study\cite{shuntingPaper} being considered has been developed by Thales
(now Ground Transportation Systems) and focuses on a train during shunting movements.
The system includes an AI-based perception system
and a deterministic steering system.
The role of the perception system is to detect
and classify obstacles (persons, animals, vehicles, ...)
and railway infrastructure elements. The steering system then
makes appropriate decisions about moving the
locomotive based on that information.

There was a set of requirements provided with the case
study, including the correct detection of several shunting train
signs. In order to increase confidence in the perception system we aim
to check the recognized signs with a runtime monitor. This will
give strong confidence that detected signs are correct. In order to safeguard
against unrecognized signs we will need to lean on other measures taken by the
project, like a thorough environment ontology
and systematic test case generation~\cite{FrauenhoferTestCases}.

\subsection{Certified Control}

Certified Control\cite{certcontrol} is an architectural framework for the real-time validation of
autonomous systems. It distinguishes itself from conventional monitoring
components by omitting its reliance on independent perception and instead counting on the
controller to provide a \emph{certificate} containing all essential information.
This certificate serves as input for the runtime monitor, which assesses the
accuracy of system behavior against specified criteria. By adopting this approach, the
architecture establishes a trusted foundation that can potentially be subjected
to a rigorous formal verification process.

The controller, which is not included in the \emph{trusted base}, can utilize
sophisticated algorithms such as neural networks without needing explicit formal
verification. By separating the tasks of generating visual insights and ensuring
safety, established verification methods can continue to be used with minimal
adjustments. To accomplish this, a formal acceptance specification for the
certificate is necessary to ensure compliance with safety requirements like
\emph{the detected lane lines are parallel} or \emph{there are no objects on the
track for 100m}. This reduces the amount of code needing verification and allows the AI
components to go unverified.

While the effectiveness of this architecture in lane line detection for regular
vehicles is promising~\cite{certcontrol}, its applicability to other autonomous perception tasks
such as sign classification and object detection remains uncertain.
Therefore, we aim to investigate the applicability and
effectiveness of such a certified control architecture in the context of the case
studies train control perception system.

\subsection{Related Work}

Other attempts at verifying an autonomous train perception systems
notably include~\cite{StandardisationConsiderations}. The authors propose a
multi-sensor pipeline relying on the statistical independence of the different
perception mechanisms to control hazards and ensure suitable model performance.
The goal is to show possible ways of certifying according to the ANSI/UL
4600~\cite{ul4600} standard, which  provides a framework for integrating AI
into fully autonomous systems. The standard gives practical guidelines and advice for a
possible safety case, notably including the entire autonomy pipeline and AI
algorithms. We also hope to provide methods to aid with a verification according
to this standard, while a full certification is currently out of reach.
Other approaches to formal runtime monitor verification of AI systems have been done in the
field of reinforcement learning using safety
shields~\cite{ShieldSynthesisForRL}. But these approaches focus on training agents to choose
optimal policies depending on given environmental factors, which is similar
to the traditional steering system in our model.
There have also been proposals for formalizing image specification, including
spatial model checking~\cite{DBLP:journals/corr/CianciaLLM16} and attempts to
formalize vision ontology~\cite{CVOntology, visionOntology}.

\section{Specification and Ontology}
\label{sec:spec}

\begin{figure}[ht]
  \vspace{-2mm}
  \centering
  \subfloat[Sh0]{\includegraphics[width=0.20\linewidth]{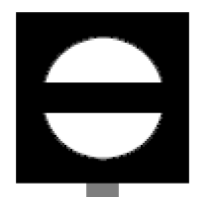}\label{fig:checking-a}}
  \hspace{.02\linewidth}
  \subfloat[Sh1]{\includegraphics[width=0.20\linewidth]{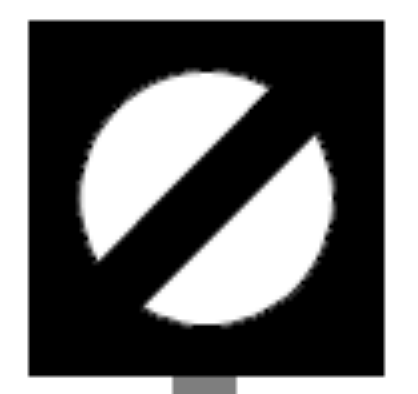}\label{fig:checking-b}}
  \hspace{.02\linewidth}
  \subfloat[Wn7]{\includegraphics[width=0.20\linewidth]{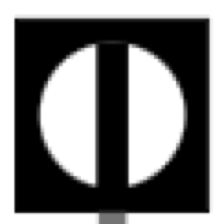}\label{fig:checking-c}}

  \caption{Train Control Shunting Signs\label{fig:signs}}
  \vspace{-2mm}
\end{figure}

The selected sign classes for verification are Sh0, Sh1, and Wn7 as depicted
in~\cref{fig:signs}. While these look similar, the semantic content is
different. Sh0 means stop and the others signal safe passage. This
makes properly distinguishing them a safety-critical issue.
To ensure that the train comes to a stop when encountering a Sh0 sign on the
current track, it is crucial to accurately detect and locate it. To
achieve this, we employ an AI object detection system in
the controller. Subsequently, the monitor verifies if the bounding box image
aligns with the expected ontology. This provides additional confidence in the
accuracy of the result.

It is often challenging to provide a precise formal definition of an image class
based solely on its features. Instead, we focus directly on detectable image characteristics. In this context, we can
observe that the images include two semi-circles with only
orientation as the distinguishing feature. This characteristic feature
allows us to define the sign using the contours and orientation angles
of the feature.

For a given image tensor $I$ with height $h$ and width $w$,
consider the set of contours (sets of points) denoted as $C(I)$, which are identified by a
contour detection algorithm. Let $S_0$ be the set of images
belonging to the Sh0 class. Also define $A: C(I) \rightarrow
\mathbb{R^+}$ as the area function, which calculates the area of a given
contour. Similarly, let $\sigma: C(I) \rightarrow \mathbb{Z}^+$ be an
orientation function that determines the angle between the contour and the
horizontal axis.
We can then express membership of an image to one of the classes by
considering an image a member of the set $S_0$ if it contains a pair $(c_1, c_2)
\in C(I) \times C(I)$, which fullfills all the following conditions, given some
pre-determined error tolerances $\delta_i, i \in \{1,2,3,4,5\}$\footnote{In the prototype
  implementation the tolerance values used were $\delta_{1,2,5} = 0.2$, $\delta_3 = 0.1$
  and $\delta_4 = 0.3$} and an expected angle $a$ that depends on the class in question.

\begin{enumerate}
\item $ A(c_1) (1-\delta_1) \leq  A(c_2) \leq (1+\delta_1) A(c_1)$
\item $(1-\delta_2)\sigma(c_1) \leq \sigma(c_2) \leq (1+\delta_2) \sigma(c_1)$
\item  $\delta_3 h \leq A(c_i) \leq \delta_4 h , i\in {1,2}$
\item  $\delta_3 w \leq A(c_i) \leq \delta_4 w , i\in {1,2}$
\item  $ c_1 \cap c_2 = \emptyset$
\item  $  |\sigma(c_i)-a| \leq 90\delta_5 , a=0$
\end{enumerate}

For the remaining two classes, the expected angle $a$ in the final condition varies
to 45 for Sh1 and 90 for Wn7. Otherwise, the definitions are identical.
The conditions one to six define an Sh0 sign as an image with
two contours that have similar angles and orientations. The orientation should
be within a certain error threshold. Also, the definition expects, that the
areas do not overlap. While ideally, we expect an orientation of
zero, variations can occur due to different photo angles. Thus, the inclusion of
an error term accounts for this discrepancy in measurement accuracy.

This definition is not flawless and permits the possibility of false
positives. This implies that there may be instances where images that do not
depict the intended sign could potentially be accepted (see \cref{fig:fpacc}). However, incorporating
this check reduces the likelihood of such occurrences compared to those without
it. The stringency of the monitoring process needs to be weighed against the
decrease in true positives to strike a suitable balance. Adjustments
can be made by selecting appropriate $\delta$ values within certain limits.
Now we can define a requirement for a correct implementation.

\emph{REC}: The implementation accurately verifies whether an image meets the ontology requirements of a specific class.

\section{Implementation and Experiments}

While the following implementation is not yet verified in terms of \emph{REC},
we aim to do so in future work. Here we provide a prototype, which is developed
enough to indicate the potential usefulness of such an implementation.
Given an image and an expected class, it either validates or rejects the image.
We then integrated it with a YOLOv8 object detection model and
measured the influence on common performance metrics (see.
\cref{table:results-with-mon}). In the following sections, we present details
on the implementation and the performed experiments.

\subsection{Implementation}

The controller component is a simple wrapper for
the YOLOv8\footnote{\url{https://github.com/ultralytics/ultralytics}} implementation
of an object detection model known as YOLO~\cite{DBLP:journals/corr/RedmonDGF15}.
The outcomes obtained from this model are packaged into a certificate and
transmitted to the monitor.
To have the model detect the signs in question, we created and
labeled a custom sign-detection dataset~\cite{sign-detection-4oqe4_dataset},
on which we trained three model variants. These were the nano, small and medium
versions of the model with 3.2M, 11.2M and 25.9M parameters respectively.
The training was done for 200 epochs with a batch size of 16.
They achieved mAP50 values of 0.827, 0.90 and 0.93 on the test set.

From the model results the controller generates a \emph{certificate} -- in the
sense of certified control (see \cref{sec:related}) -- consisting of the following components:
\begin{enumerate}
\item The original image.
\item The assigned class result.
\item The bounding box, represented as a tuple in the format $(x,y,w,h)$, with values normalized to fit the dimensions of the image.
\end{enumerate}
This Python object is then given to the monitor. In a production
implementation, it would be preferable to serialize and send this data to a
statically typed version of the monitor for optimal security.

The  monitor implementation utilizes Python's
OpenCV~\cite{itseez2015opencv} library to apply simple and well-tested
computer vision algorithms to the given images.
To begin, the bounding box image is resized to 206x206 and converted to
grayscale to facilitate contour detection. Subsequently, a filtering process is
applied to the contours to ensure their area falls within the specified size
boundaries (refer to \cref{sec:spec}). We then need to
calculate the area and orientation of the detected contours to
determine if some of them fit the requirements for the ontology.
The area of each contour is extracted using an available function within OpenCV.
In addition, we utilize OpenCV once again by fitting a line through each contour
as a means of determining its orientation.
With that, we can calculate the orientation using the following equation.
$$ \sigma(c) = \frac{180\arccos (\vec{e_1} \cdot \vec{v})}{\pi|\vec{v}|} $$
Next, we evaluate the remaining contours in pairs to determine if they satisfy
the similarity conditions for area and orientation (refer to \cref{sec:spec} for details). If a pair is found that
meets these conditions, we then verify if its orientation aligns with the
expected orientation for the corresponding class. If it does, the monitoring
system considers this as a valid certificate. However, if any of these criteria
are not met, the certificate will be rejected.

\begin{figure}[ht]
  \vspace{-2mm}
  \centering
  \subfloat[FP Sh0 accepted]{\includegraphics[width=0.20\linewidth]{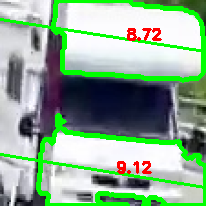}\label{fig:fpacc}}
  \hspace{.025\linewidth}
  \subfloat[FP Sh0 rejected]{\includegraphics[width=0.20\linewidth]{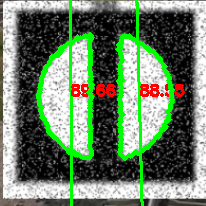}\label{fig:fprej}}
  \hspace{.025\linewidth}
  \subfloat[TP Sh0 accepted]{\includegraphics[width=0.20\linewidth]{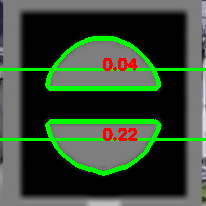}\label{fig:tpacc}}
  \hspace{.025\linewidth}
  \subfloat[TP Sh0 rejected]{\includegraphics[width=0.20\linewidth]{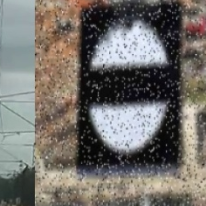}\label{fig:tprej}}

  \caption{Visual Examples of Successful and Failing Monitor Checks\label{fig:monitor}}
  \vspace{-2mm}
\end{figure}

\subsection{Experiments}

In contrast to the automotive field, which benefits from large-scale image
datasets like KITTI~\cite{KITTI} for efficient object detection model evolution using road
scene images, the railway industry faces limitations in terms of relevant
datasets. Recently, interesting multi-sensor benchmark datasets~\cite{OpenSensorDataForRail} have
started to emerge, but do not fit our particular use case.
This lack of labeled, high-quality data poses a challenge when it comes to training and
validating AI-based systems for this particular case.
When evaluating the performance of the prototype, we have to confront this
lack of data in the field. Since the relevant publicly available datasets do not cover
the classes in question, we resort to custom labeling for training and a data
generation approach for the evaluation of the system.
For the generation, we chose a small number of base images of the signs in question,
which are put through different random perturbation combinations and
then pasted in random amounts -- one to four -- onto images from train
footplate rides, gathered from the web.
By this method we generated 28283 unique images containing 43638 signs. There
are up to four signal per picture, which is typical of a shunting yard. For this
work we ignore the selection of relevant signals and only focus on detection.
The following perturbations were applied:
\begin{enumerate}
\item Horizontal Flip
\item Gaussian Noise (Salt and Pepper with Levels of 0.05 and 0.075)
\item Scaling (Up and back down to square images of 50,100,213,416 and 832 px)
\item Blur (normalized box filter with kernel sizes 3, 5, 7)
\item Brightness change (levels 0.5,1.5)
\end{enumerate}
In \cref{fig:monitor} we see examples of these images with monitor
visualizations applied. It shows cut YOLO bounding boxes with the contours,
lines and corresponding orientations detected by the monitor. \cref{fig:fpacc}
shows one of the few remaining false positives. The image fits all the defined
criteria of the $S_0$ ontology for these $\delta$ values but is not actually of
that class. Given stricter tolerances (e.g. $90\delta_5 < 8$) this mistake would
not occur. Overall the results seen in~\cref{table:results-with-mon} show a slight reduction in
model performance in terms of recall and an evenly weighted F-score compared to
the prior results in~\cref{table:results-without-mon}. The concrete detection
numbers can be found in~\cref{table:numbers}.
The drop in recall and F-score is expected due to the reduction in true positives.
However, almost all false positives have been recognized and can thus be
prevented. The tolerances can be adjusted to further reduce false positives, at the cost
of more recall and F-score, or to allow more leeway to the perception system.
In terms of runtime performance, the monitor checks a certificate in
approximately 0.7 ms on an Intel i5-12600K processor.
In comparison, the inference of the YOLOv8 model will range from 2 ms -- for the
nano model variant -- to 8 ms for the m version. This means that the
performance overhead is likely not a major concern in a production environment.

\begin{table}
  \vspace{-5mm}
  \centering
  \subfloat[Results without Monitor]{%
    \begin{tabular}{lrrr}
      Model & Detected & TP & FP\\[0pt]
      \hline
      n & 30111 & 25514 & 4597\\[0pt]
      s & 30335 & 26790 & 3545\\[0pt]
      m & 28672 & 22728 & 5944\\[0pt]
    \end{tabular}
    \label{table:numbers-without-mon}
  }\quad
  \subfloat[Results with Monitor]{%
    \begin{tabular}{lrrr}
      Model & Detected & TP & FP\\[0pt]
      \hline
      n & 21716 & 21714 & 2\\[0pt]
      s & 22834 & 22831 & 3\\[0pt]
      m & 20460 & 20460 & 0\\[0pt]
    \end{tabular}
    \label{table:numbers-with-mon}
  }
  \caption{Raw numbers for Models on Generated Data}
  \label{table:numbers}
  \vspace{-5mm}
\end{table}

\begin{table}
  \centering

  \subfloat[Results without Monitor]{%
    \begin{tabular}{lrrr}
      Model & Precision & Recall & $F_1$ score\\[0pt]
      \hline
      n & 0.85 & 0.58 & 0.69\\[0pt]
      s & 0.88 & 0.61 & 0.72\\[0pt]
      m & 0.79 & 0.52 & 0.63\\[0pt]
    \end{tabular}
    \label{table:results-without-mon}
  }\quad
  \subfloat[Results with Monitor]{%
    \begin{tabular}{lrrr}
      Model & Precision & Recall & $F_1$ score\\[0pt]
      \hline
      n & 1.00 & 0.50 & 0.67\\[0pt]
      s & 1.00 & 0.52 & 0.68\\[0pt]
      m & 1.00 & 0.47 & 0.64\\[0pt]
    \end{tabular}
    \label{table:results-with-mon}
  }

  \caption{Model Metrics on Generated Data (values rounded to two decimal places)}
\end{table}

\section{Conclusion and Future Work}
\label{sec:conclusion}

In conclusion, this study demonstrates the potential utility of certified
control runtime monitoring for object detection of formally definable and safety
critical classes. The resulting trade-off in our tests is promising enough
to warrant further investigation into different application possibilities.
However, further research is necessary to fully validate its
implementation in a type-safe language following the \emph{REC} guidelines.
The obtained results should be verified in
appropriate field test for any real world application.
Additionally, it should be noted that a significant portion of the perception
system remains unverified.
Moving forward, our future work will involve evaluating the applicability of a
similar architecture for other components of the perception system such as
obstacle detection. This evaluation will include examining different sensor
types such as LIDAR and radar on benchmark datasets.

\nocite{*}
\bibliographystyle{eptcs}
\bibliography{paper}

\end{document}